\documentclass{article}

\PassOptionsToPackage{numbers, compress}{natbib}
\usepackage[preprint]{neurips_2024}

\usepackage{caption}
\usepackage{amsmath} 

\usepackage{graphicx}
\usepackage{multirow}
\usepackage[utf8]{inputenc} % allow utf-8 input
\usepackage[T1]{fontenc}    % use 8-bit T1 fonts
\usepackage{hyperref}       % hyperlinks
\usepackage{url}            % simple URL typesetting
\usepackage{booktabs}       % professional-quality tables
\usepackage{amsfonts}       % blackboard math symbols
\usepackage{nicefrac}       % compact symbols for 1/2, etc.
\usepackage{microtype}      % microtypography
\usepackage{xcolor}         % colors
\hypersetup{
    colorlinks=true,
    linkcolor=blue,
    citecolor=blue,
    urlcolor=blue
}
\usepackage{algorithm}
\usepackage{algorithmic}
\title{Improved GUI Grounding via Iterative Narrowing}

\author{%
  Anthony Nguyen \\
  Algoma University\\
  Ontario, Canada \\
  \texttt{anguyen@algomau.ca} \\
}
\begin{document}

\maketitle

% TODO: Emphasis on no need to finetune. But still comparable results to other works that did. As finetuning requires a lot of computation and training data.
\begin{abstract}
Graphical User Interface (GUI) grounding plays a crucial role in enhancing the capabilities of Vision-Language Model (VLM) agents. While general VLMs, such as GPT-4V, demonstrate strong performance across various tasks, their proficiency in GUI grounding remains suboptimal. Recent studies have focused on fine-tuning these models specifically for zero-shot GUI grounding, yielding significant improvements over baseline performance. We introduce a visual prompting framework that employs an iterative narrowing mechanism to further improve the performance of both base and finetuned models in GUI grounding. For evaluation, we tested our method on a comprehensive benchmark comprising various UI platforms and provided the code to reproduce our results in this \href{https://github.com/ant-8/GUI-Grounding-via-Iterative-Narrowing}{GitHub repository}.
\end{abstract}

\section{Introduction}

GUI Grounding is the task of identifying the visual location on an interface image given a natural language query~\cite{cheng2024seeclick, wu2024atlas}. This task is essential for improving the capabilities of Vision Language Model (VLM) agents, enabling them to better understand and interact with graphical user interfaces~\cite{OSWorld}. While general VLMs, such as GPT-4V, have demonstrated high performance across a variety of visual-linguistic tasks, their accuracy in GUI grounding tasks remains suboptimal~\citep{cheng2024seeclick, OSWorld}.

Existing solutions have attempted to bridge this performance gap by fine-tuning VLMs for GUI grounding, resulting in improvements over baseline models~\citep{gou2024uground, cheng2024seeclick, wu2024atlas, hong2023cogagentvisuallanguagemodel}. However, these approaches often come with the need for extensive data and re-training.

Instead of training a model, we propose a visual prompting framework that employs an iterative narrowing mechanism to improve the grounding performance of current VLMs. Our approach treats the model's initial position predictions as an approximation. The framework iteratively refines the prediction by focusing on progressively smaller, cropped regions that center around the previous prediction. This refinement process can be repeated for multiple iterations.

\begin{figure}[h]
    \centering
    \includegraphics[width=0.99\textwidth]{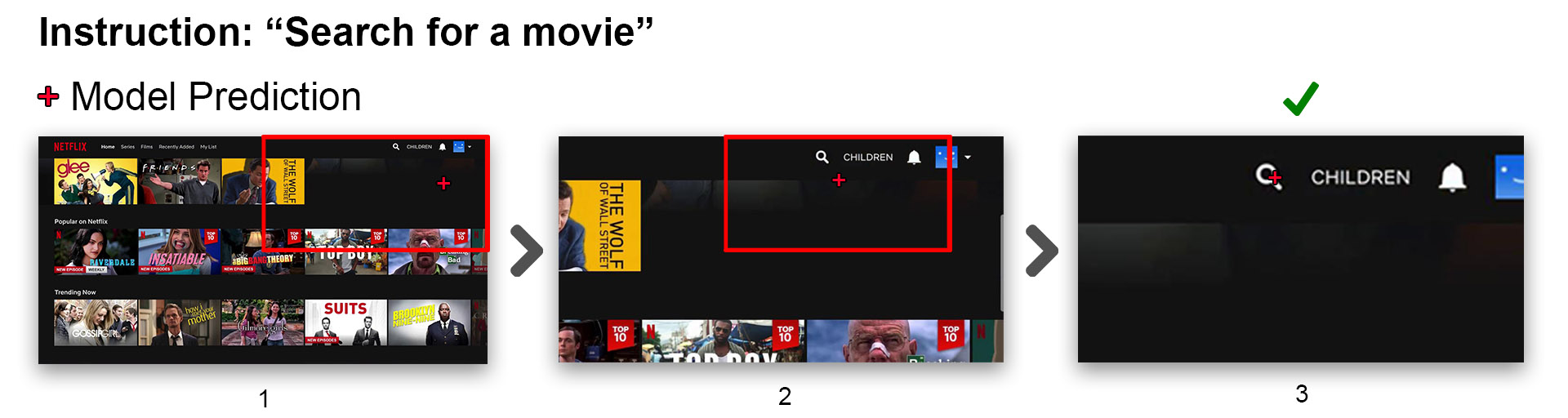} % 
    \caption{A visual demonstration of our method. On each iteration, the model prediction (red cross) determines the next region (red box) to iterate.}
    \label{fig:demonstration}
\end{figure}

We evaluate the performance of our method on the ScreenSpot~\citep{cheng2024seeclick} benchmark. Our findings demonstrate that this iterative approach leads to substantial performance gains, providing a straightforward yet effective solution for enhancing GUI grounding capabilities.

To encourage reproducibility and facilitate further research, we provide our code at \url{https://github.com/ant-8/GUI-Grounding-via-Iterative-Narrowing}.

\section{Related Work}
\label{headings}

\subsection{Iterative Reasoning and Localization}
Iterative refinement strategies have been explored in multiple  contexts, aiming to progressively narrow down the target location. 
Wu et al.~\cite{wu2023vguidedvisualsearch} proposed SEAL, an LLM-guided visual search framework designed for multimodal large language models in high-resolution natural image understanding, particularly for visual question answering tasks. 
SEAL integrates a dedicated visual search model, specialized localization decoders, and a Visual Working Memory to actively recover missing visual details, requiring additional model inference beyond the base VLM. 
In contrast, our iterative narrowing method is developed specifically for precise GUI element grounding, is training-free, and reuses the same VLM across successive refinement steps. 
Chen et al.~\cite{chen2018iterativevisualreasoningconvolutions} introduced a visual reasoning framework that combines spatial memory with graph reasoning to iteratively refine predictions through spatial and semantic cues. 
Similarly, Sun et al.~\cite{sun2021iterativeshrinkingreferringexpression} proposed an iterative shrinking approach for referring expression grounding and relies on reinforcement learning to determine the shrinking strategy, whereas we perform refinement purely through repeated model predictions without additional training or policy learning.

\subsection{Grounding Models}
Recent work on GUI grounding has explored both large-scale pretraining and task-specific model design. 
Cheng et al.~\citep{cheng2024seeclick} trained SeeClick, a finetuned grounding model, and introduced the ScreenSpot benchmark, highlighting the benefits of grounding-specific pretraining. 
Gou et al.~\citep{gou2024uground} achieved up to 20\% accuracy gains using a pixel-level model trained on 10 million GUI elements. 
Wu et al.~\citep{wu2024atlas} developed OS-Atlas models with a 13-million GUI element dataset, delivering consistent improvements across multiple benchmarks. 
Lin et al.~\citep{lin2024showuivisionlanguageactionmodelgui} proposed ShowUI, a lightweight vision-language-action model optimized for GUI tasks through efficient token selection and streaming methods.

\subsection{GUI Agents}
Beyond grounding in static scenarios, other works have explored GUI agents that perform end-to-end interaction. 
Xie et al.~\cite{OSWorld} introduced the OSWorld benchmark to evaluate such agents, emphasizing grounding in screenshot-only settings. 
Wu et al.~\cite{wu2024atlas} improved task performance by combining OS-Atlas-Base-7B with GPT-4o ~\cite{openai2024gpt4ocard}  for enhanced reasoning and perception. 
Hong et al.~\cite{hong2023cogagentvisuallanguagemodel} presented CogAgent, which serves as both a grounding model and an interactive visual agent.

\section{Methodology}

Our method refines the prediction of target coordinates within an input image \(\mathcal{I}\), guided by a text query \(\mathcal{Q}\), through an iterative process of cropping and coordinate prediction.

Initially, the entire image \(\mathcal{I}\) is treated as the cropping window, with dimensions equal to the original image size. At each iteration, the model predicts normalized coordinates \((x_k, y_k) \in [0, 1] \times [0, 1]\), representing a relevant location within the current cropping window. These coordinates are used to update the cropping window for the next iteration, progressively narrowing the region of focus.

The cropping window dimensions are adjusted dynamically based on the image orientation. For landscape-oriented images, the width and height are scaled by factors \(\alpha_{\text{landscape}}\) and \(\beta_{\text{landscape}}\), respectively, where both scale factors are less than one. Similarly, for portrait-oriented images, the width is scaled by \(\alpha_{\text{portrait}}\), which is less aggressive than the landscape scaling, while the height is scaled by \(\beta_{\text{portrait}}\).

After \(n\) iterations, the final normalized coordinates are transformed into absolute coordinates relative to the original image dimensions. This transformation ensures that the output \((X, Y)\) corresponds directly to a location within the original image.

\subsection{Pseudocode}

The following pseudocode outlines the iterative process of predicting the target location in the image:

\begin{algorithm}[H]
\caption{Iterative Cropping and Prediction}
\textbf{Input:} Image \(\mathcal{I}\), Query \(\mathcal{Q}\), Number of iterations \(n\) \\
\textbf{Output:} Final coordinates \((X, Y)\) relative to the original image
\begin{enumerate}
    \item Initialize cropping window \(\mathcal{C}_0\) dimensions \(w_0 = W_{\text{orig}}, h_0 = H_{\text{orig}}\)
    \item For \(k = 1\) to \(n\):
    \begin{enumerate}
        \item Using VLM: Predict normalized coordinates \((x_k, y_k) \in [0, 1] \times [0, 1]\) relative to \(\mathcal{C}_{k-1}\), from \(\mathcal{Q}\)
        \item If \(\mathcal{I}\) is landscape-oriented:
        \begin{itemize}
            \item \(w_k \gets \alpha_{\text{landscape}} \cdot w_{k-1}, \quad h_k \gets \beta_{\text{landscape}} \cdot h_{k-1}\)
        \end{itemize}
        \item Else (portrait-oriented):
        \begin{itemize}
            \item \(w_k \gets \alpha_{\text{portrait}} \cdot w_{k-1}, \quad h_k \gets \beta_{\text{portrait}} \cdot h_{k-1}\)
        \end{itemize}
        \item Define cropping window \(\mathcal{C}_k\):
        \begin{itemize}
            \item Center: \((x_k, y_k)\)
            \item Dimensions: \((w_k, h_k)\)
        \end{itemize}
    \end{enumerate}
    \item Convert normalized \((x_n, y_n)\) to coordinates relative to the original image space \((X, Y)\)
    \item \textbf{Return} Final coordinates \((X, Y)\)
\end{enumerate}
\end{algorithm}

\section{Experiments}

\subsection{Setup}
We evaluated our method on the ScreenSpot~\citep{cheng2024seeclick} benchmark, a comprehensive assessment tool for single-step GUI grounding across multiple platforms. The ScreenSpot benchmark is structured into three primary categories: \textit{mobile}, \textit{web}, and \textit{desktop}, each designed to reflect common user interface environments. Furthermore, each category is divided into two subcategories based on the type of target element: \textit{text} or \textit{icon}.

For the entire evaluation, we used \( n = 3 \) iterations. While this choice provided promising results, we did not conduct in-depth investigations to determine an optimal value for \( n \) or how aggressive each crop should be shrunken. The cropping aggressiveness was controlled using scaling factors:

\begin{itemize}
    \item For landscape-oriented images:
    \begin{itemize}
        \item Width scaling factor: \(\alpha_{\text{landscape}} = 0.5\)
        \item Height scaling factor: \(\beta_{\text{landscape}} = 0.5\)
    \end{itemize}
    \item For portrait-oriented images:
    \begin{itemize}
        \item Width scaling factor: \(\alpha_{\text{portrait}} = \frac{1}{1.2} \approx 0.833\)
        \item Height scaling factor: \(\beta_{\text{portrait}} = 0.5\)
    \end{itemize}
\end{itemize}

Future work may explore the impact of varying \( n \) and the values of \(\alpha_{\text{landscape}}\), \(\beta_{\text{landscape}}\), \(\alpha_{\text{portrait}}\), and \(\beta_{\text{portrait}}\) on performance and inference time.

\newpage
\subsection{Evaluation Results}

\begin{table}[h!]
    \centering
    \begin{tabular}{llcc}
        \toprule
        \textbf{Model Type} & \textbf{Models} & \textbf{Baseline} & \textbf{IN} \\
        \midrule
        \multirow{2}{*}{Base} & InternVL-2-4B & 4.32 & \textbf{6.53} \\
                                 & Qwen2-VL-7B & 42.89 & \textbf{69.1} \\
        \midrule
        \multirow{2}{*}{Finetuned} & ShowUI-2B & 75.1 & \textbf{79.56} \\
                                       & OS-Atlas-Base-7B & 82.47 & \textbf{83.33} \\
        \bottomrule
    \end{tabular}
    \vspace{0.15cm}
    \caption{Overall average accuracy (\%) comparing baseline against our method (IN) on the ScreenSpot benchmark.}
    \label{tab:overall_accuracy}
\end{table}

\begin{table}[h!]
    \centering
    \begin{tabular}{llcccc}
        \toprule
        \textbf{Model Type} & \textbf{Models} & \multicolumn{2}{c}{\textbf{Text}} & \multicolumn{2}{c}{\textbf{Icon/Widget}} \\
        \cmidrule(lr){3-4} \cmidrule(lr){5-6}
        & & Baseline & IN & Baseline & IN \\
        \midrule
        \multirow{2}{*}{Base} & InternVL-2-4B & 9.16 & \textbf{14.65} & 1.64 & \textbf{2.18} \\
                                   & Qwen2-VL-7B & 61.34 & \textbf{83.52} & 39.29 & \textbf{57.21} \\
        \midrule
        \multirow{2}{*}{Finetuned} & ShowUI-2B & \textbf{92.3} & 89.01 & \textbf{75.5} & 74.24 \\
                                       & OS-Atlas-Base-7B & 93.04 & 93.04 & 72.93 & \textbf{74.24} \\
        \bottomrule
    \end{tabular}
    \label{tab:mobile_scores}
    \vspace{0.15cm}
        \caption{Accuracy scores in the \textit{mobile} category.}
\end{table}

\begin{table}[h!]
    \centering
    \begin{tabular}{llcccc}
        \toprule
        \textbf{Model Type} & \textbf{Models} & \multicolumn{2}{c}{\textbf{Text}} & \multicolumn{2}{c}{\textbf{Icon/Widget}} \\
        \cmidrule(lr){3-4} \cmidrule(lr){5-6}
        & & Baseline & IN & Baseline & IN \\
        \midrule
        \multirow{2}{*}{Base} & InternVL-2-4B & 4.64 & \textbf{8.76} & \textbf{4.29} & 2.14 \\
                                   & Qwen2-VL-7B & 52.01 & \textbf{84.54} & 44.98 & \textbf{60.71} \\
        \midrule
        \multirow{2}{*}{Finetuned} & ShowUI-2B & 76.3 & \textbf{87.6} & 61.1 & \textbf{68.57} \\
                                       & OS-Atlas-Base-7B & 91.75 & \textbf{92.27} & 62.86 & \textbf{77.14} \\
        \bottomrule
    \end{tabular}
    \label{tab:desktop_scores}
    \vspace{0.15cm}
        \caption{Accuracy scores in the \textit{desktop} category.}
\end{table}

\begin{table}[h!]
    \centering
    \begin{tabular}{llcccc}
        \toprule
        \textbf{Model Type} & \textbf{Models} & \multicolumn{2}{c}{\textbf{Text}} & \multicolumn{2}{c}{\textbf{Icon/Widget}} \\
        \cmidrule(lr){3-4} \cmidrule(lr){5-6}
        & & Baseline & IN & Baseline & IN \\
        \midrule
        \multirow{2}{*}{Base} & InternVL-2-4B & 0.87 & \textbf{5.22} & 0.10 & \textbf{2.91} \\
                                   & Qwen2-VL-7B & 33.04 & \textbf{73.04} & 21.84 & \textbf{50.0} \\
        \midrule
        \multirow{2}{*}{Finetuned} & ShowUI-2B & 81.7 & \textbf{83.04} & 63.6 & \textbf{68.93} \\
                                       & OS-Atlas-Base-7B & \textbf{90.87} & 86.96 & \textbf{74.27} & 72.33 \\
        \bottomrule
    \end{tabular}
    \label{tab:web_scores}
    \vspace{0.15cm}
        \caption{Accuracy scores in the \textit{web} category.}
\end{table}

% TODO: 
\subsection{Weaknesses \& Observations}

\begin{figure}[h]
    \centering
    \includegraphics[width=0.96\textwidth]{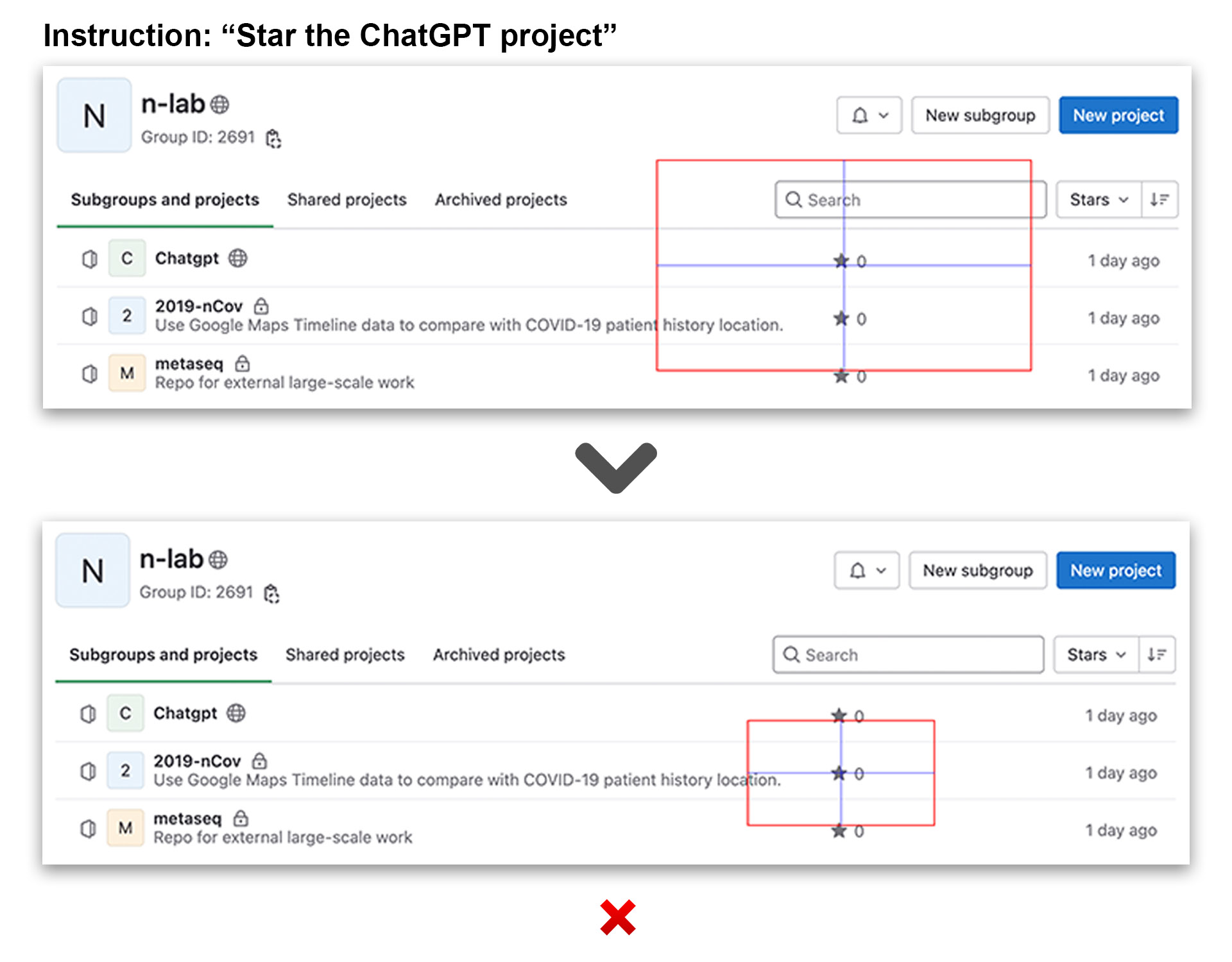} % 
    \caption{An example of a context-dependent failure case. At iteration \( k < n \) (top image), the cropping leads to the loss of crucial contextual information. Specifically, the association between each star button and its corresponding project. Thus, resulting in the wrong UI element being selected (bottom image).}
    
    \label{fig:context_dependant_case}
\end{figure}

A key weakness of our method is the loss of contextual information as iterations progress, posing challenges for identifying target elements that rely on spatially distant cues (see Figure~\ref{fig:context_dependant_case}).

Our method improves precision but at the cost of underperforming in context-dependent scenarios. This limitation likely explains the greater gains observed in generalist VLMs (e.g., InternVL-2-4B~\cite{chen2024far, chen2023internvl}, Qwen2-VL-7B~\cite{Qwen2VL}) versus OS-Atlas~\cite{wu2024atlas} and ShowUI~\cite{lin2024showuivisionlanguageactionmodelgui}, which already has strong precision abilities from training. Addressing this limitation by maintaining context throughout the iterations could further enhance performance for all models.

\subsection{Potential Solutions and Future Work}

To address the loss of contextual information, we conducted preliminary experiments where the model was provided with both the entire screenshot and the current crop during each iteration. The goal was to enable the model to retain a global understanding while refining its focus through local crops.

However, during these initial experiments, we observed that the VLM frequently confused the local crop with the global context image. This confusion led to incorrect coordinate predictions, as the model sometimes generated coordinates referencing the entire global image rather than accurately focusing on the local context intended for refinement.

We believe that further fine-tuning could help the model more effectively distinguish between global and local contexts, ultimately improving its performance in context-dependent scenarios.

\section{Conclusion}
In this work, we introduced a visual prompting framework designed to perform an iterative narrowing mechanism. As a result, this enhances the GUI grounding capabilities of Vision-Language Models (VLMs). Through iterative refinement of predictions, our method improves accuracy by narrowing the focus on progressively smaller regions, allowing models to more precisely identify visual elements on a graphical user interface. Evaluation on the ScreenSpot~\cite{cheng2024seeclick} benchmark demonstrated that our method improves performance, especially for generalist VLMs like InternVL-2-4B~\cite{chen2023internvl,chen2024far} and Qwen2-VL-7B~\cite{Qwen2VL}, where substantial improvements were observed compared to baselines. However, the method shows limitations when handling spatially distant contextual cues, which affects performance in context-dependent scenarios.

Future work may focus on addressing these contextual limitations by incorporating both global and local context information more effectively. Promising directions may include refining the model's ability to differentiate between local crops and the entire image. There is potential to further push the boundaries of VLM precision in GUI grounding tasks, contributing to the development of more effective visual agents.

\bibliographystyle{plainnat} % Choose your preferred bibliography style
\bibliography{references} % This should match the name of your .bib file

\end{document}